\documentclass[sigconf, nonacm, screen]{acmart}
\usepackage{longtable}
\usepackage{graphicx}
\usepackage[ruled]{algorithm}
\usepackage{algorithmic}
\usepackage{amssymb}
\usepackage{amsmath}
\usepackage{bm}
\usepackage{subcaption}
\usepackage[flushleft]{threeparttable}

\newcommand{\argmin}{\mathop{\rm arg~min}\limits}

\AtBeginDocument{%
  \providecommand\BibTeX{{%
    \normalfont B\kern-0.5em{\scshape i\kern-0.25em b}\kern-0.8em\TeX}}}

\begin{document}

\title{Unbiased Lift-based Bidding System}

\let\SUP\textsuperscript
\author{Daisuke Moriwaki}
  \affiliation{%
  \institution{CyberAgent, Inc.}
}
\email{moriwaki\_daisuke@cyberagent.co.jp}

\author{Yuta Hayakawa}
  \affiliation{%
  \institution{CyberAgent, Inc.}
}
\email{hayakawa\_yuta@cyberagent.co.jp}

\author{Isshu Munemasa}
  \affiliation{%
  \institution{CyberAgent, Inc.}
}
\email{munemasa\_isshu@cyberagent.co.jp}

 \author{Yuta Saito} 
    \affiliation{%
 \institution{Tokyo Institute of Technology}
}
\email{saito.y.bj@m.titech.ac.jp}

\author{Akira Matsui} 
    \affiliation{%
  \institution{University of Southern California}
  }
\email{amatsui@usc.edu}

\renewcommand{\shortauthors}{Moriwaki et al.}

\begin{abstract}
Conventional bidding strategies for online display ad auction heavily relies on observed performance indicators such as clicks or conversions. A bidding strategy naively pursuing these easily observable metrics, however, fails to optimize the profitability of the advertisers. Rather, the bidding strategy that leads to the maximum revenue is a strategy pursuing the performance \textit{lift} of showing ads to a specific user. Therefore, it is essential to predict the lift-effect of showing ads to each user on their target variables from observed log data. However, there is a difficulty in predicting the lift-effect, as the training data gathered by a past bidding strategy may have a strong bias towards the winning impressions. In this study, we develop \textit{Unbiased Lift-based Bidding System}, which maximizes the advertisers' profit by accurately predicting the lift-effect from biased log data. Our system is the first to enable high-performing lift-based bidding strategy by theoretically alleviating the inherent bias in the log. Real-world, large-scale A/B testing successfully demonstrates the superiority and practicability of the proposed system.
\end{abstract}



\maketitle

\section{Introduction}
\label{sec: introduction}
Online advertising has been playing an integral role in the recent business, which accounts for half of media ad spending in the world~\cite{noauthor_global_nodate}. One of the advantages of the online ad is that the performance of the ads can be assessed by simple metrics such as the number of clicks and conversions earned by the ads. These metrics enable ad deliverers, DSPs (Demand-Side Platforms), to charge ad costs for advertisers through the objective billing systems named CPC (cost-per-click), CPA (cost-per-action), and CPM (cost-per-impression). The conventional goal of DSPs is to optimize these metrics by performance-based bidding strategy, which decides bid price based on the probability of users to take the desired action.

Despite its industrial success, there are two major emerging issues with the current bidding process.
First, the conventional bidding strategy ignores the probability that a user will convert even without an ad. Such a strategy is suboptimal in the sense that it will not reach users having a high probability of \textit{changing} their action by showing an ad~\cite{xu2016lift}. Moreover, it might destroy the experience of end-users who have high conversion probability even without an ad~\cite{moriwaki_fatigue-aware_2020}.
The other problematic issue is the inherent bias in training data.
Specifically, for supervised machine learning to work, it is necessary for training and test population to follow the same distribution.
However, this is not the case in the online advertising domain.
This is because the training data is heavily biased by the ad auction selection in the data collection process (i.e., \textit{impression bias}).
Impressions that were assigned to a higher bid price by a past bid strategy would have a higher probability of winning in an auction, thereby having a higher chance to appear in the training data.
On the other hand, one has to make a prediction for every impression before the ad auction selection in the testing time.
As a result, machine learning models trained on that biased data will overly focus on samples with a high winning
probability, resulting in a poor performance in the testing time~\cite{zhang2016bid}.

These two issues have been solved separately. \cite{xu2016lift} shows that the conventional strategy optimizing observed clicks or conversions is suboptimal and proposes a lift-based bidding strategy. This strategy decides the bid price based on the predicted \textit{lift} effect of showing an ad to a user and considers their conversion probability without the ad. However, their proposed method does not address the impression bias theoretically. In contrast, \cite{zhang2016bid} proposes a method to alleviate the bias and unbiasedly predict the outcome under the impression bias. However, the bidding strategy considered in~\cite{zhang2016bid} still pursues the observed metrics, not the lift-effect. Therefore, a theoretically grounded method that simultaneously solves these two challenging problems has not yet been proposed in the literature.

In this study, we solve the aforementioned issues simultaneously with the aim of maximizing advertisers' revenue while avoiding damaging end-users' experience.
We first propose a method to predict the lift of showing ads to a specific user using only an observable biased impression log.
Our method builds on a theoretically validated debiasing method in causal inference called \textit{inverse propensity scoring} and easily implementable with existing machine learning packages such as \textit{scikit-learn}\footnote{https://scikit-learn.org/stable/} or \textit{XGBoost}\footnote{https://xgboost.readthedocs.io/en/latest/}.
We further describe the details of our resulting \textit{Unbiased Lift-based Bidding System} architecture used in our real production.
Our system is scalable and able to optimize advertisers' profit by deciding the bid price for each impression based on our proposed unbiased lift prediction procedure.
Finally, we conduct online A/B testing and demonstrate that our proposed system outperforms the naive, conventional performance-based bidding strategy in a real ad campaign.

Our contributions can be summarized as follows:
\begin{itemize}
\item We propose a theoretically grounded, easily implementable method to predict the lift-effect of showing ads from biased impression log data.
\item We develop \textit{Unbiased Lift-based Bidding System}, which maximizes the advertisers' profit by bidding based on the predicted lift-effect.
\item We demonstrate the efficacy of our proposed system in a large-scale online A/B testing on a real-world DSP.
\end{itemize}

\section{Motivation}
This study aims at increasing the effectiveness of an advertising campaign by optimizing bid prices under budget constraints. To explain this, we provide a brief overview of the bidding problem.

Marketers craft a whole picture of their advertising campaign and split a fraction of the budget for online ads, spending the rest to other advertisements such as TVs and newspapers. With that specific assigned budget, DSPs join online ad auctions. This pre-determined constraint is almost always an equity constraint and the budget DSPs want to consume without any excess and deficiency. Hence the tasks of DSPs are twofold: \textbf{1)} allocating the campaign budget for online advertisements efficiently to eat up its budget and \textbf{2)} acquiring an effective impression in  RTB (Real-time Bidding).

DSPs are unable to know which or when, even how many users show up in the auction, but only able to predict them. Hence, DSP needs to determine the appropriate bid price for each bid request sequentially by learning the quantity and quality of bid requests to keep budget constraints. Then the DSP's problem at $k$th bid request is to choose the best bid price $bid_k \in \mathbb{R}_{>0}$ to maximize expected profit over $[k, \bar{k})$ period. Let $wp_k$ be the winning price of the auction and $\omega_k$ be the indicator function that tells if the bidder wins or not. $wp_k$ is the highest bid for first price auction and second highest bid for second price auction. $\omega_k$ takes one if the DSP wins and takes zero if not. We treat them as a function of $bid_k$.

Let $v_j$ be the {\it value} obtained from the ad-slot $j$. The conventional performance-based bidding strategy uses the probability of conversion with an ad as the value as follows.
\begin{align}
v_j = E[y_i \mid ad_j] .
\label{eq:performance_based_value} 
\end{align}

It just uses the observed performance of the ad, but ignores the action changes caused by the ad. In contrast, the lift-based bidding strategy set $v_j$ as the {\it lift} of conversions caused by the ad impression. Namely, 
\begin{align}
v_j = E[y_i \mid ad_j] - E[y_i \mid no \, ad_j].
\label{eq:value} 
\end{align}
Then the DSP solves the following maximization problem.
\begin{align}\label{eq:opt_prob}
& \max\limits_{bid_k} \ E\left[\sum\limits_{j \in [k, \bar{k})} \left( v_{j}\omega_j(bid_j) - wp_j(bid_{j})\omega_j(bid_j) \right)\right] 
\\ 
& \quad {\rm s.t. } \ E\left[\sum\limits_{j \in [k, \bar{k})}  wp_{j}(bid_j)\omega_{j}(bid_j)\right] = B_k, 
\end{align}
where the second equation implies that the expected ad inventory cost $\sum\limits_{j \in [k, \bar{k})} wp_j(bid_{j})\omega_j(bid_{j})$ must meet remaining budget at $k$, $B_k$.\footnote{In the real-world production the formulation of the constraint depends on advertisers' interest. $wp$ can be cost-per-click or cost-per-impression.}
With the above optimization problem, DSPs should gain an effective impression having a large lift-effect to maximize the profit. Our focus in this study is to propose a method to predict the lift-effect with the biased impression data and implement a system that controls the bid prices based on the predicted lift-effect.

\section{Background and Related Work}\label{sec:problem_back_set}
We briefly summarize the background of this study and critical related works.

DSPs participate in the online auctions to purchase ad impressions from SSP (Supply-Side Platform). In general, DSPs charges advertisers for their performance measured by an observed metric such as the number of clicks or conversions. Specifically, they claim a fixed price for each click (cost-per-click, CPC), or each conversion (cost-per-action, CPA). As a result, performance-based bidding strategies aim at maximizing the number of clicks or conversions after impression (Eq.~(\ref{eq:performance_based_value})). 

Just focusing on CPC or CPA, however, may yield suboptimal strategy because they omit the probability that the targeted users get converted (e.g., visit a store, purchase a good) without advertising. To incorporate this omitted realm, the DSP needs to predict incremental gain caused by ad impression (Eq.~(\ref{eq:value})).

The attempts to incorporate this lift-effect in bidding strategy are lift-based bidding~\cite{xu2016lift}, incrementality bidding~\cite{johnson2015cost}, or more broadly uplift modeling~\cite{kawanaka2019,zaniewicz2013support,radcliffe2011real,saito2019doubly,saito2020cost}. We advance this line of literature and address the inherent bias in impression log data in a theoretically grounded and tractable manner. Specifically, we propose a method to unbiasedly predict the lift-effect of an ad impression on a specific user from biased impression data. Moreover, our method is easily implementable with the well-known machine learning libraries, while the previous debiasing method for the performance-based bidding strategy needs additional implementation cost~\cite{zhang2016bid}. We believe that our method will broaden the application of the promising lift-based bidding strategy. This is because every online ad system is biased by the ad auction selection~\cite{zhang2016bid}, and practitioners can implement and try our unbiased lift-effect prediction method immediately. To our knowledge, we are the first to theoretically consider both the lift-effect prediction and ad impression bias in the online advertising literature.

\section{Proposed Method}
\label{sec: proposed method}
In this section, we propose and describe our \textit{Unbiased Lift-based Bidding System}.

\subsection{Setup}
We consider a bidding strategy that rewards $v$ for each lift $\tau$ so that a DSP earns $v$ for each extra conversion (as in Eq.~(\ref{eq:value})). Our algorithm calculates the bid price $bid_{i}$ as:
\begin{eqnarray}\label{eq:bidding}
    bid_{i,s(a)} &=& \alpha \cdot v \cdot \tau(s(a) \mid \mathbf{x}_i).
\end{eqnarray}
where $\alpha \in (0,1)$ is cost rate, a budget pace multiplier that will be detailed in Section~\ref{sec:cost-rate-parameter}. $v$ is some fixed value a DSP earns for conversion lift. In our algorithm, $\tau$ is the predicted lift-effect of advertising $a$ for user $i$ characterized by a feature vector $\mathbf{x}_i$. $s(a) \in \mathcal{S}$ represents the ad exposure state of an ad $a$ under consideration, and $\mathcal{S}$ is a set of possible states. We use the number of impressions of $a$ to $i$ as a scalar variable representing the ad exposure state of $a$ to $i$, and thus $\mathcal{S} = \{0, 1, \ldots \}$ here.

To formally define the lift-effect $\tau$, we introduce the essential notation called \textit{potential outcomes} in causal inference~\cite{imbens2015causal}. Let $y_i (s(a))$ denote user $i$'s potential outcome associated with the exposure state $s(a)$ of ad $a$. Each user $i$ has potential outcomes associated with every possible state, i.e., $ \{ y (s(a)) \ | \ \forall s(a) \in S \}$, however, only one of them is observable. The observed outcome for user $i$ is defined as $ y_i^{obs} =  y_i (s_i)$ where $s_i$ is a realized exposure state for user $i$. Note that the potential outcomes associated with every possible state other than the realized one, i.e., $\{ y (s(a)) \ | \ \forall s(a) \in S \backslash \{s_i\} \}$ is unobservable or counterfactual to the analysts.

The lift-effect $\tau$ of showing ad $a$ for each user $i$ is sequentially defined as the difference between the expectation of the potential outcomes given the two consecutive ad exposure states (the number of impressions):
\begin{align}\label{eq:tau}
\tau(s (a) \ | \ \mathbf{x}_i) 
& = E[ y_i (s(a) \ | \ \mathbf{x}_i ] -  E[ y_i (s(a) - 1) \ |\ \mathbf{x}_i] , \forall s(a) \in \mathcal{S} \backslash \{0\}.
\end{align}
where $ E[ y_i (s(a)) |\mathbf{x}_i  ] $ is the expected potential outcome of $i$ when the number of impressions is $s(a)$, in contrast $ E[ y_i (s(a) - 1) |\mathbf{x}_i] $ is the expected potential outcome when the number of impressions is $s(a)-1$. Thus, Eq.~(\ref{eq:tau}) is the reasonable definition for the lift-effect of showing an ad $a$ one more time to a specific user $i$ who has been exposed to the ad $s(a)-1$ times in the past.

To predict $\tau$, we train predictors of outcomes for every possible state $\forall s \in \mathcal{S}$ separately and combine their predictions as follows. 
\begin{align}\label{eq:tau_f}
\hat{\tau}(s(a) \ | \ \mathbf{x}_i) = f^{s(a) }(\mathbf{x}_i) - f^{s(a)-1}(\mathbf{x}_i), \forall s(a) \in \mathcal{S}\backslash \{0\}.
\end{align}
where $f^{s(a)}$ and $f^{s(a)-1}$ predict $E[ y_i (s) \ | \ \mathbf{x}_i] $ and $E[ y_i (s(a)-1) \ | \ \mathbf{x}_i]$, respectively. To accurately predict the lift-effect $\tau$, it is essential to predict the expected probability of conversion under each ad exposure state well.

\subsection{Unbiased Lift-effect Prediction}
\label{subsec:attribution}
To obtain a well-performing predictor $f$ for each ad exposure state, it is ideal to directly optimize the following generalization error:
\begin{align}
    \mathcal{L}_{ideal} (f^{s(a)}) = E_{p(x,y)} [ \ell (y (s(a)), f^{s(a)} (\mathbf{x})) ], \forall s(a) \in \mathcal{S}.
\label{eq:ideal_loss}
\end{align}
where $f^{s(a)}$ is a predictor for $ E[y (s(a)) \ | \ \mathbf{x}] $, $\ell$ specifies a loss function such as mean squared error, $p(x,y)$ is joint probability distribution of the \textbf{whole population}, meaning that the population before the ad auction selection or the testing time. We consider optimizing the generalization error defined over the whole population because we apply $f^{s(a)}$ to predict the potential outcome in the testing time. 

In reality, however, it is impossible to directly optimize Eq.~(\ref{eq:ideal_loss}). This is because we can utilize only a finite size $n_{s(a)}$ of training data $\mathcal{D}_{s(a)} = \{ (\mathbf{x}_i, y_i^{obs}) \ |\ s(a)=s_i\}_{i=1}^{n_{s(a)}} \sim p(x,y  | s(a))$ for each ad state and cannot take the expectation to obtain Eq.~(\ref{eq:ideal_loss}). The conventional solution to this is \textit{Empirical Risk Minimization} (ERM), which optimizes the empirical approximation of Eq.~(\ref{eq:ideal_loss}) as: 
$$f^{s(a)}_{ERM} = \argmin_{f^{s(a)}} \hat{\mathcal{L}}_{ERM} (f^{s(a)}) =  \argmin_{f^{s(a)}}  \frac{1}{n_{s(a)}} \sum_{i \in \mathcal{D}_{s(a)}} \ell (y_i^{obs}, f^{s(a)} (\mathbf{x}_i))$$.

The ERM principal works well under the situation of the same train-test distribution, however, the ad impression bias breaks this premise of machine learning. Specifically, the simple empirical approximation of the loss function over $\mathcal{D}_{s(a)}$ has a bias, i.e., $ E_{p(x,y,s(a))} [\hat{\mathcal{L}}_{ERM} (f^{s(a)})] \neq \mathcal{L}_{ideal} (f^{s(a)})$ for some given $f^{s(a)}$.
The bias issue emerges because users assigned to a higher bid price has higher density in the training data than that in the test data (i.e., $p(x,y) \neq p (x,y | s(a))$). As a result, the trained predictor $ f^{s(a)}_{ERM}$ may perform poorly in the testing time, because it mistakenly overfits to the over-represented samples in the training data.

To alleviate this bias issue of ERM in the online adverting situation, we apply the \textit{inverse propensity score} (IPS) estimation technique to debias the estimation of the ideal loss in Eq.~(\ref{eq:ideal_loss}). Our loss function takes the following form:
\begin{align}
    \hat{\mathcal{L}}_{IPS} (f^{s(a)}) =  \frac{1}{n} \sum_{i \in \mathcal{D}_{s(a)}} \frac{1}{e_{s_i} (\mathbf{x}_i)} \ell (y_i^{obs}, f^{s(a)} (\mathbf{x}_i)).
\label{eq:ips_loss}
\end{align}
where $n = \sum_{s(a) \in \mathcal{S}} n_{s(a)}$ is the total number of the training data, and $ e_{s_i} (\mathbf{x}_i) = P (s(a) = s_i | \mathbf{x}_i ) $ is the probability of user $i$ being assigned to the ad exposure state $s_i$ called the \textit{propensity score}. A fascinating property of the IPS loss in Eq.~(\ref{eq:ips_loss}) is that it is unbiased for the ideal generalization error as the following proposition states.
\begin{proposition} The IPS loss function in Eq.~(\ref{eq:ips_loss}) is unbiased for the ideal generalization error in Eq.~(\ref{eq:ideal_loss}), i.e., for any given $f^{s(a)}$, we have
\begin{align*}
    E_{p(x,y,s(a))} [ \hat{\mathcal{L}}_{IPS} (f^{s(a)}) ] = \mathcal{L}_{ideal} (f^{s(a)})
\end{align*}
under standard identification assumptions in causal inference~\cite{imbens2015causal,saito_unbiased_2020,saito2019doubly}.
\end{proposition}

The above proposition suggests that our IPS loss function successfully alleviates the bias issue of ERM and approximates the ideal loss from only observable data. Therefore, to unbiasedly predict the lift-effect under the ad impression bias, we optimize the IPS loss and use the resulting predictors to obtain the final lift-effect prediction as:
\begin{align*}
    \hat{\tau}(s(a) \ | \ \mathbf{x}_i) = f_{IPS}^{s(a)}(\mathbf{x}_i) - f_{IPS}^{s(a)-1}(\mathbf{x}_i), \forall s(a) \in \mathcal{S} \backslash \{0\}.
\end{align*}
where $ f_{IPS}^{s(a)} =  \argmin_{f^{s(a)}} \hat{\mathcal{L}}_{IPS} (f^{s(a)})$ is the IPS loss minimizer.

\subsection{PID Control for Cost Rate}
\label{sec:cost-rate-parameter}
We use $\alpha$ for budget pacing, and it discounts the predicted lift-effect in the auctions as in Eq.~(\ref{eq:bidding}). Appropriate $\alpha$ is not known a priori, and thus we use PID (Proportional Integral Derivative) control algorithm following~\cite{zhang2016feedback} to adaptively update it to keep the spending constant. We tune the parameters $k_p, k_i, k_d$, and parametric specification of the actuator in Algorithm 1 by an offline simulation. We found that the performance is stable when using the exponential actuator, i.e., $exp(\cdot)$ as a parametric specification.

\begin{algorithm}[htb]
\caption{PID control for cost rate}         
\begin{algorithmic}                  
\REQUIRE Default cost rate $\alpha^\ast$, hyperparameters $k_p, k_i, k_d$
\STATE Set $\alpha$ = $\alpha^\ast$
\WHILE{$h < H$}
\STATE $err_h$ = remaining budget/remaining hours - budget spending at $h-1$
\STATE $\alpha_{h+1} = exp \left( err_h k_p + \sum\limits_{j \le h}err_j k_i + (err_h - err_{h-1}) k_d \right)  \alpha_h$
\ENDWHILE
\end{algorithmic}
\label{alg:PID} 
\end{algorithm}

\section{Implementation}
\label{sec: system design and implementation}
We implement the proposed algorithm in a unique DSP specialized for online-to-offline marketing provided by CyberAgent, Inc, a Japan-based ad agency. Our DSP uses location data to determine bid prices and treat users' visits to offline stores as conversions.

\subsection{Architecture}
\label{subsec:architecture}
\begin{figure}[htb]
  \centering
  \includegraphics[width=0.8\linewidth]{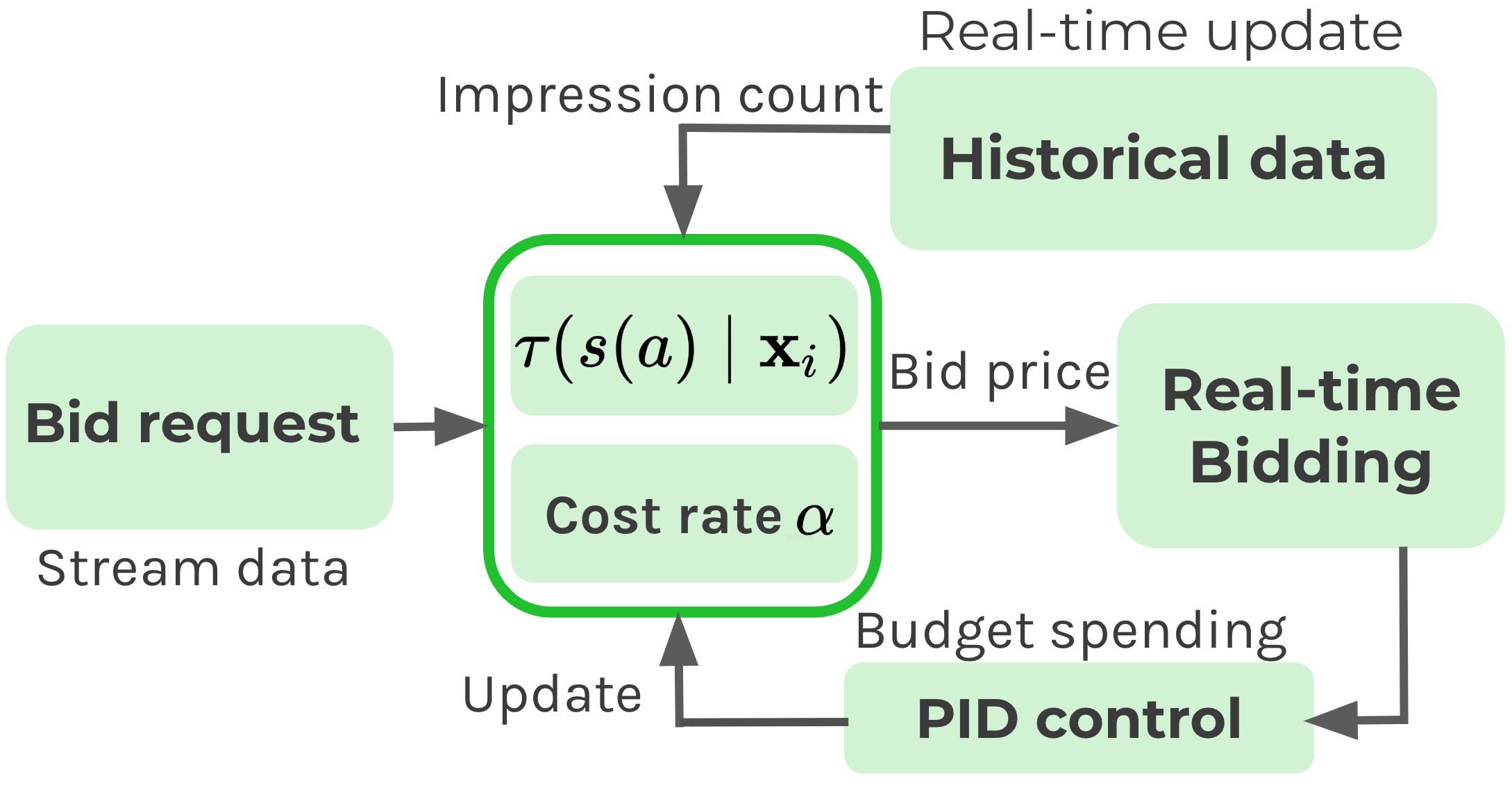}
  \caption{The lift-based bidding system architecture}  \label{fig:system}
\vskip 0.03in
\begin{minipage}{\columnwidth} 
{\small {\it Note}: For each bid request, the DSP server confirms how many impressions the coming user has been exposed to the ad in the past (impression count). The PID controller updates the cost rate $\alpha$ every hour based on the gap between the ideal and realized budget spendings. Then DSP calculates the bid price by combining the cost rate $\alpha$ and pre-predicted lift-effect of exposing the ad to the user one more time $\tau$.}
\end{minipage}
\end{figure}

We summarize the whole architecture of our bidding system in Figure~\ref{fig:system}. For each bid request, corresponding $\tau$ multiplied by cost rate $\alpha$ is returned as the bid price. $\tau$ is a predicted lift-effect for a coming impression. The ad impression count is calculated by scanning the history of an ad impression for each user. The PID controller updates the cost rate $\alpha$ in every hour based on the gap between ideal budget spending and realized budget spending. The whole procedure is scalable and done within a few milliseconds and does not harm the user experience. 

\subsection{Preprocessing of Impressions}
\begin{figure}[htb]
  \centering
  \includegraphics[width=0.9\linewidth]{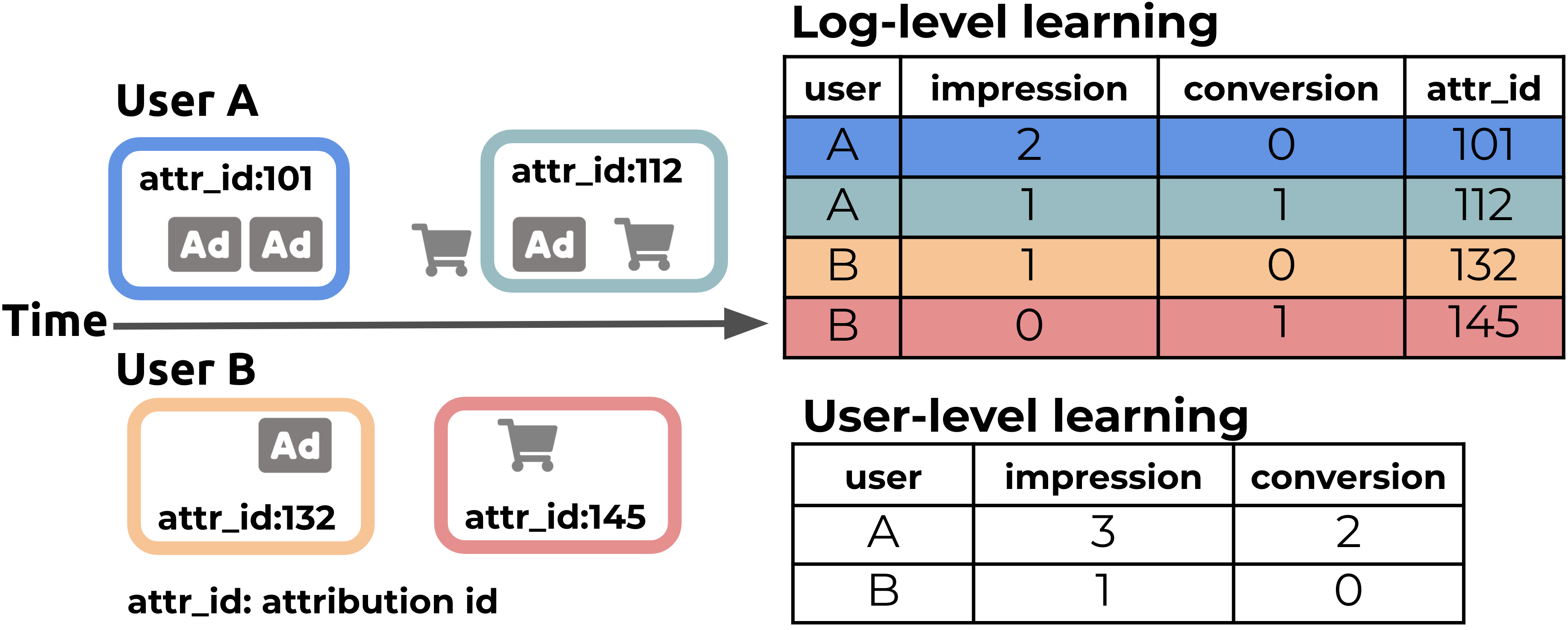}
  \caption{Comparing log-level and user-level learning}
\label{fig:user level learning}
\vskip 0.03in
\begin{minipage}{\columnwidth} 
{\small {\it Note}: The log-level learning captures the higher granularity logs than the user-level learning. The log-level learning samples the attributions from the users' timeline giving multiple attribution ids (attr\_ids) to the users, and the user-level learning summarizes the whole history of the timeline for each user. The table of log-level learning can have multiple rows (i.e., multiple attr\_ids) for a single user, whereas the table of user-level learning has one row per user.
\par}
\end{minipage}
\end{figure}

To attribute each conversion to ad impressions, we use {\it user-level learning} rather than {\it log-level learning} used in existing research. Log-level learning implies randomly sampling data points from users' timeline by windows (the rectangles in Figure~\ref{fig:user level learning}). The method used in \cite{xu2016lift} can be classified into this group. On the other hand, user-level learning firstly composes user level data and learns the relationship between impressions and conversions for each user. Note that the data size of user-level learning is the number of users, whereas log-level learning generates larger data.

User-level learning has several advantages over log-level learning. First, it does not require any hyperparameter for data splitting such as window size. Second, user-level learning incorporates whole data points including those omitted in log-level learning. One drawback of user-level learning is that the data size is small. This, however, poses little problem for our setting, because DSPs usually reaches out to millions of users for each campaign.

\subsection{Lift-effect Model Training}
\label{subsec: model training}

\noindent
\textbf{Data Sources}.
We take the training data from two types of sources. When there are similar advertising campaigns in the past, we use the data for training. When no similar advertising campaign is conducted before we introduce lift-based bidding in the middle of the ongoing campaign and use the data generated at that point. \\

\noindent
\textbf{Ad sizes}.
The online ad takes various forms, including creative, template, format, and size. Size is especially critical as it is standardized by IAB\footnote{\url{https://www.iab.com/}}. To account for the effect of ad size on target variables, Essentially, we group ads into four by size and train size-specific predictors.
\\

\noindent
\textbf{Propensity Score Estimation}.
To use our IPS loss function, we have to estimate the propensity score $e_{s}(\cdot)$ in Eq.~(\ref{eq:ips_loss}). The estimation of the propensity score is the multi-class classification problem. We use multi-class classifier from XGBoost library~\cite{chen_xgboost:_2016} and train it using the whole training data $\{( \mathbf{x}_i, s_i )\}_{i=1}^n$ and classifies the ad exposure state $s_i$ from the feature vectors. The feature vector to estimate the propensity score includes estimated CVR used for the existing bidder and the number of impressions before the advertising campaign of training period as well as the geographical features of user $i$. The estimated CVR and the number of impressions before the advertising campaign improve the estimation quality of the propensity score because the higher estimated CVR means higher bid price and the number of impressions in the training period indicates how easy the ad-slots of the user is obtained. For each ad size, we split users into eight classes by the number of impressions $s_i \in \mathcal{S}=\{0, 1, ,2 ,3, 4, 5-9, 11-20, 20+\}$. The window size of bins gets larger for the higher number because the distribution of the number of impressions is highly skewed to the right. We train four propensity score predictors for each ad size. \\

\noindent
\textbf{Outcome Prediction}.
We train outcome predictors for each pair of ad exposure states and ad sizes. We use XGBoost regressor, since the outcome variable $y$ is the number of users' visits to an offline store. To train a regressor, we optimize the IPS loss function in Eq.~(\ref{eq:ips_loss}). XGBoost provides a weighting option, thus our IPS loss minimization procedure is easily implementable. The feature vector $\mathbf{x}$ includes recency (the last time the user visited the stores), frequency (how many times the user visited the stores), visited POIs (point-of-interests), the number of location logs, census statistics of the user's residential areas such as the share of age groups, averaged size of the households, and land prices. 

\subsection{The flooring and smoothing $\tau$}
The predicted lift-effect can be negative.  While it is natural for the lift-effect prediction framework to predict the negative effect of the ad, bidding negative price is not rational and could cause a problem in the system that does not expect negative bid price. Hence we first set $\tau$ to zero for $\tau < 0$. The floored values are still very volatile across the impression count. It is possible that the expected value of the second impression is very high, while the first impression has zero expected value. Then the impression count for the user never gets two because the bidding for the first impression always loses. To deal with the problem, we smoothed the values with 3-impression count backward-moving average so that the value of the first impression gets high enough in the above example.

\section{Online Experiment}
\label{sec: the online experiment}

\subsection{Experimental Design}
To evaluate our unbiased lift-based bidding system, we compared its performance indicators (impressions-per-user, etc.), business KPIs (key performance indicators), and its pacing efficiency with a conventional bidding system through an online A/B testing. The performance-based system employs the community standard bidding system, which decides bid price by Eq.~(\ref{eq:performance_based_value}). Note that our lift-based bidding system determines the bid price by the predicted lift-effect of showing one additional ad to a user and is described in Eq.~(\ref{eq:value}) and Eq.~(\ref{eq:bidding}). In the A/B testing, we randomly assigned the two systems to users on an online ad campaign that aims to promotes the app released by a major consumer electronics retailer in Japan\footnote{
The experiment duration was June 9-14 2020, where COVID-19 had have been a serious issue across the world. Although we understand that the COVID-19 could influence our experiment, we conclude that it did not pose serious issues. First of all, we conducted after the Japanese government called off the state of emergency. Also, the state of emergency even depends voluntarily rather than compulsory, and, unlike other several countries, strict measurements such as ``lockdown" were not taken. 
}. The primary aim of the campaign is to increase the number of app users and visitors to the real stores located across Japan. To alleviate the impacts of the experiment on the business, we make a relatively small group for the lift-based group, which results in the unbalanced experiment groups. However, we keep a sufficiently large group size for statistical analysis. 
\subsection{Results and Discussion}

\begin{table}[t]\small
\centering
\caption{A/B testing; Lift-based vs Performance-based}
\begin{tabular}{lccl}\toprule
    & {\small Lift-based}& {\small Performance-based} & {\small Difference}\\ \midrule
\#Impressions-per-user    &  1.28 &     1.0 &  0.28*** \\
\#Clicks-per-user&  0.54 &     1.0 & -0.46** \\
Reach rate        &  1.71 &     1.0 &  0.71*** \\
\#Visit-per-user     &  1.01 &     1.0 &  0.0093 \\
Share of visitors     &  1.004 &     1.0 &  0.0036 \\\#Users & 467,180 & 3,310,182&\\
    \toprule
\end{tabular}
\label{table:results} 
\vskip 0.03in
\begin{minipage}{\columnwidth} 
{\small {\it Note}: The lift-based bidding system obtains more impressions and achieves larger reach rate than the performance-based bidding system. We divided each metric by the results of the performance-based strategy for the normalization purpose. \textbf{\#Impressions-per-user} is the average number of impressions by users; \textbf{\#Clicks-per-user} is the average number of clicks by users; \textbf{Reach rate} is the ratio of the users who got one or more impressions; \textbf{Share of visitor} is the ratio of the users who visited one or more offline stores.; The number of stars represents the statistical significance level (p-value $< 0.05$*, p-value $< 0.01$**, p-value $< 0.001$***).
\par}
\end{minipage}
\end{table}

\begin{table}[t]\small
\centering
 \caption{Business KPIs; Lift-based vs Performance-based}
\begin{tabular}{lcc}
\toprule
 &  Lift-based  &  Performance-based \\
\midrule
Cost-per-impression   &  0.27 &   1.0  \\
Cost-per-reach & 0.24 & 1.0  \\
Cost-per-visit & 0.30  &   1.0  \\

\bottomrule
\end{tabular}
\vskip 0.03in
\begin{minipage}{\columnwidth} 
{\small {\it Note}: The proposed lift-based strategy efficiently won the impressions and conversions compared to the performance-based one, and thus has positive impacts on our business. We divided each metric by the results of the performance-based strategy for the normalization purpose. The table shows the three essential business KPIs. \textbf{Cost-per-impression} is the number of impressions per dollar spent for the ad inventories; \textbf{Cost-per-reach} is the number of users with one or more impressions per dollar; \textbf{Cost-per-visit} is the number of visits per dollar. 
}
\end{minipage}
\label{tab:business}
\end{table}

\begin{figure}[t]
\includegraphics[width=\columnwidth]{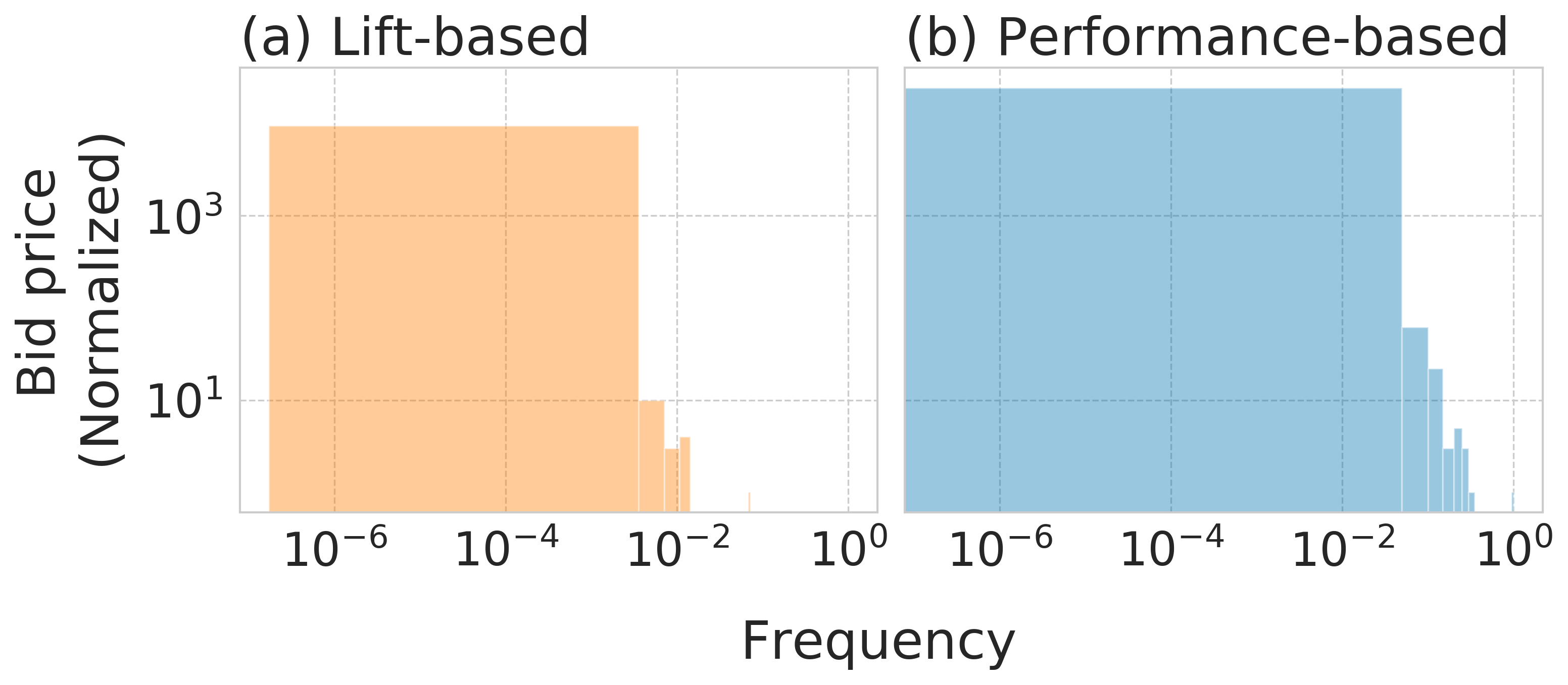}
\caption{The bid prices; Lift-based vs performance-based}
\label{fig:dist_bitprice}
\begin{minipage}{\columnwidth} 
\vskip 0.03in
{\small {\it Note}: This figure compares the distribution of the bidding price by \textbf{(a)~Lift-based bidding system} and \textbf{(b)~Performance-based bidding system}, showing that the bid prices by lift-based strategy are much smaller ($\approx10^{-1}$ times) than those by the performance-based. We computed the mean bid price for each user and min-max normalized the values where the highest value in the population takes 1, and the lowest value takes 0.\par}
\end{minipage}

\end{figure}

\begin{figure}[t]
\includegraphics[width=0.88\linewidth, height=0.65\linewidth]{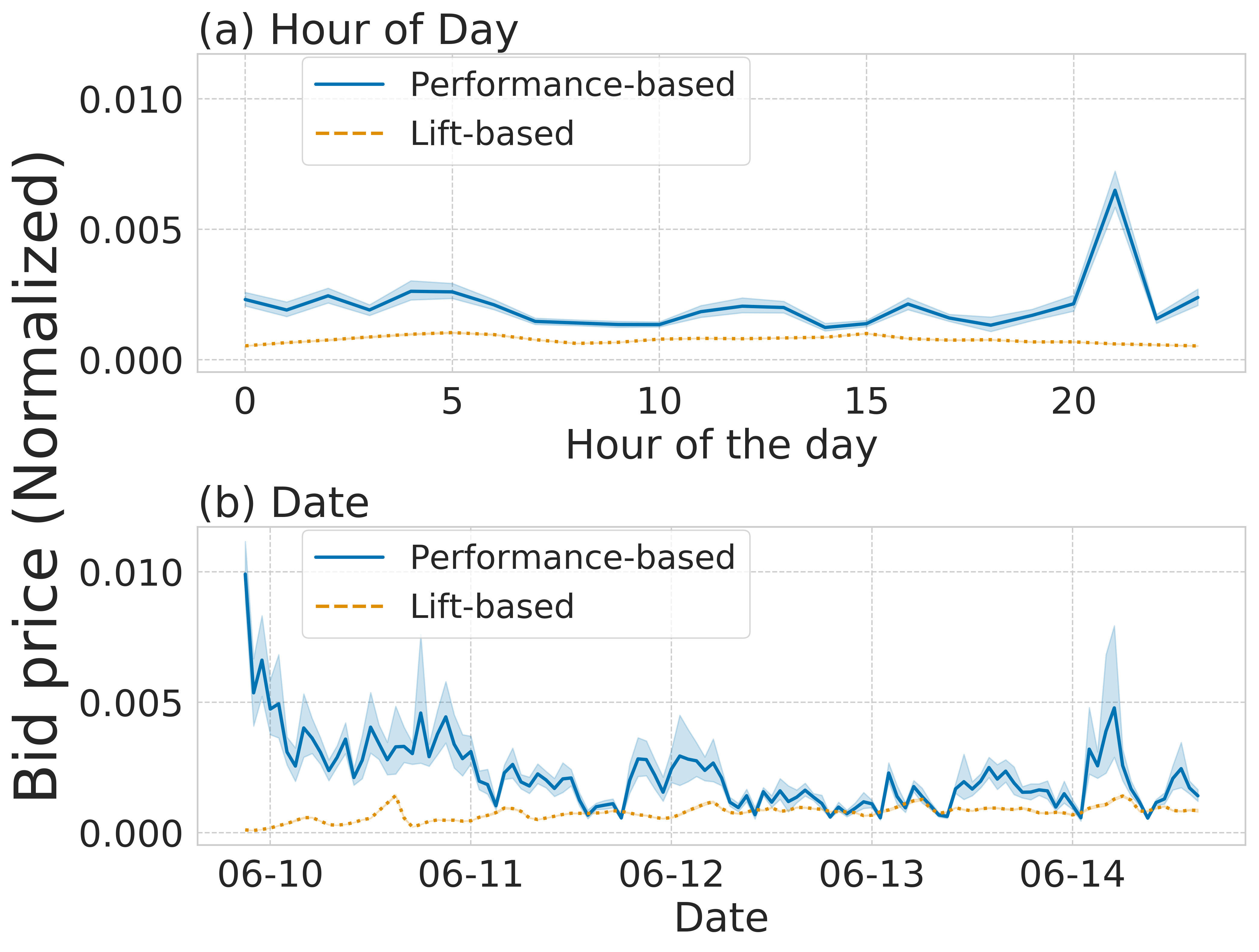}
\caption{Bid price transitions during the experiment.}   \label{fig:bid_trans}
\begin{minipage}{\columnwidth} 
\vskip 0.03in
{\small {\it Note}: This figure compares the bid price transition by \textbf{Lift-based bidding system} (solid blue line) and \textbf{Performance-based bidding system} (orange dot line) over the course of the experiments on two different time scale: \textbf{(a) Hour of day} and \textbf{(b) Date}. In both time scales, the lift-based bidding system bids a much smaller price with smaller fluctuations than the performance-based bidding system. We first compute the mean bid price for each user in each hour in each time scale. Then, we min-max normalize the data where the highest value in the population takes 1, and the lowest value takes 0. The shaded area represents 95\% confidence interval.\par}
\end{minipage}

\end{figure}

We summarize the results of the A/B testing in  Table~\ref{table:results}. Compared to the performance-based bidding system, the lift-based bidding system achieved a higher impressions-per-user (28\% more) and a higher reach rate (71\% more) but obtained fewer clicks (46\% less). Consequently, the lift-based bidding system invited more visitors to the real stores on average (0.9\% more) and reached a higher share of visitors (0.4\% more). In other words, the lift-based bidding system successfully reached out to potential customers and encouraged their visits to the real stores. The lift-based system demonstrates its superiority also in the essential key business indicators (KPIs) described in Table~\ref{tab:business}. The lift-based strategy won the impressions in an extremely economical way, obtaining each impression, reaching and conversion at 24-30\% of the ad inventory cost of the performance-based. Figure~\ref{fig:dist_bitprice} plots the distribution of the bidding prices by each bidding system and highlights that the lift-based bidding system bids a much lower price than the performance-based bidding system. The observations above suggest that the lift-based bidding system (i) bid lower prices and/or (ii) win auctions at a lower price.

Moreover, our lift-based system excels at its efficiency. The lift-based bidding system bids prices with lower fluctuation than the performance-based bidding system. Figure~\ref{fig:bid_trans} depicts the transition of the average bidding prices by the two bidding systems on the two different time scales: Hour of Day and Date. In both time scales, the performance-based bidding system is much more volatile than the lift-based bidding system. This difference stems from its competitors. Because the most rival DSPs adopt performance-based algorithm, the performance-based bidding system had to compete against many competitors for the similar ad slots, suffering from low win rates. When the performance-based system could not win the auctions, its pacing parameter $\alpha$ automatically increased and fluctuated bidding prices. In contrast, the lift-based bidding system did not have many competitors in the auctions and kept the pacing parameter $\alpha$ lower, achieving relatively higher win rates.

\section{Conclusion}
In this study, we developed \textit{Unbiased Lift-based Bidding System}, which maximizes the advertisers' profit by accurately predicting the lift-effect under the impression bias. A key feature of our proposed system is \textit{unbiased lift-effect prediction}: we enabled to unbiasedly predict the lift-effect of an additional ad impression using training data biased by a past bidding strategy. We also describe our detailed implementation and system architecture to achieve the lift-based bidding system in practice. Through online A/B testing, we demonstrated the scalability and advantages of our proposed system over the conventional performance-based one.

\bibliographystyle{ACM-Reference-Format}
\bibliography{uplift-based_bidding}

\end{document}